\documentclass[runningheads]{llncs}
\usepackage[subpreambles=true]{standalone}
\usepackage[utf8]{inputenc}
\usepackage[english]{babel}
\usepackage{import}

\begin{document}


\pagestyle{headings}
\mainmatter
\def\ECCVSubNumber{100}  

\title{Chained-Tracker: Chaining Paired Attentive Regression Results for End-to-End Joint Multiple-Object Detection and Tracking
} %

\titlerunning{Chained-Tracker}
%
\author{Jinlong Peng\inst{1} $^\star$ \and
Changan Wang\inst{1}
\thanks{\textit{Equal contribution.}}
\and
Fangbin Wan\inst{2} \and
Yang Wu\inst{3} 
\thanks{\textit{Corresponding author: Yang Wu (wuyang0321@gmail.com)}}
\and
Yabiao Wang\inst{1} \and
Ying Tai\inst{1} \and
Chengjie Wang\inst{1} \and
Jilin Li\inst{1} \and
Feiyue Huang\inst{1} \and
Yanwei Fu\inst{2}
}

\authorrunning{J. Peng et al.}
%
\institute{Tencent Youtu Lab \quad
\email{\{jeromepeng, changanwang, caseywang, yingtai, jasoncjwang, jerolinli, garyhuang\}@tencent.com}
\and Fudan University \quad
\email{\{fbwan18, yanweifu\}@fudan.edu.cn}
\and
Nara Institute of Science and Technology \quad
\email{yangwu@rsc.naist.jp}
}
\maketitle

\begin{abstract}
Existing Multiple-Object Tracking (MOT) methods either follow the tracking-by-detection paradigm to conduct object detection, feature extraction and data association separately, or have two of the three subtasks integrated to form a partially end-to-end solution. Going beyond these sub-optimal frameworks, we propose a simple online model named Chained-Tracker (CTracker), which naturally integrates all the three subtasks into an end-to-end solution (the first as far as we know). It chains paired bounding boxes regression results estimated from overlapping nodes, of which each node covers two adjacent frames. The paired regression is made attentive by object-attention (brought by a detection module) and identity-attention (ensured by an ID verification module). The two major novelties: chained structure and paired attentive regression, make CTracker simple, fast and effective, setting new MOTA records on MOT16 and MOT17 challenge datasets (67.6 and 66.6, respectively), without relying on any extra training data. The source code of CTracker can be found at: \url{github.com/pjl1995/CTracker}. 

\end{abstract}

\section{Introduction}

\yang{Video-based} scene understanding and human behavior analysis are \yang{important high-level tasks} in \wca{computer vision} with many valuable \CRwfb{applications in real scene}. \blfootnote{\CRyang{This work was supported by a MSRA Collaborative Research 2019 Grant.}}They \yang{rely} on many \yang{other tasks, within} which \textit{Multiple-Object Tracking \yang{(MOT)}} is \yang{a significant} one. \wca{However, \yang{MOT} remains \CRyang{challenging} due to the existence of occlusions,} object trajectory \CRwfb{overlap}, possibly challenging background, \textit{etc.}, especially for \yang{crowded} scenes.

\yang{Despite the great efforts and encouraging progress in the past years, there are two major problems of existing \CRwfb{MOT} solutions.} 
\CRyang{One is that} most methods are based on \CRyang{the} tracking-by-detection paradigm \cite{tracking-by-detection_ICCV09}, \yang{which is plausible but suboptimal due to the infeasibility of global (end-to-end) optimization. It usually contains three sequential subtasks: object detection, feature extraction and data association.} 
However, splitting the whole task into isolated \CRyang{subtasks may lead to} local optima \CRyang{and more computation cost} than end-to-end solutions. \wca{Moreover,} data association \yang{heavily relies} on the quality of object detection\CRyang{, which by itself is hard to generate reliable and stable results across frames as it discards the temporal relationships of adjacent frames}.
    
\CRyang{The other problem is that recent MOT methods get more and more complex as they try to gain better performances.}
\emph{Re-identification} and \emph{attention} are two major points found to be helpful for improving \CRwfb{the performance of MOT}. Re-identification (or ID verification) is used to extract more robust \CRwfb{features} for data association. Attention helps the model \CRyang{to be more focused, avoiding} the distraction by irrelevant yet confusing information (e.g. the complex background). Despite their effectiveness, the involvement of them in existing solutions greatly increases the model complexity and computational cost.

\begin{figure*}[t]
\centering
\includegraphics[width=1\columnwidth]{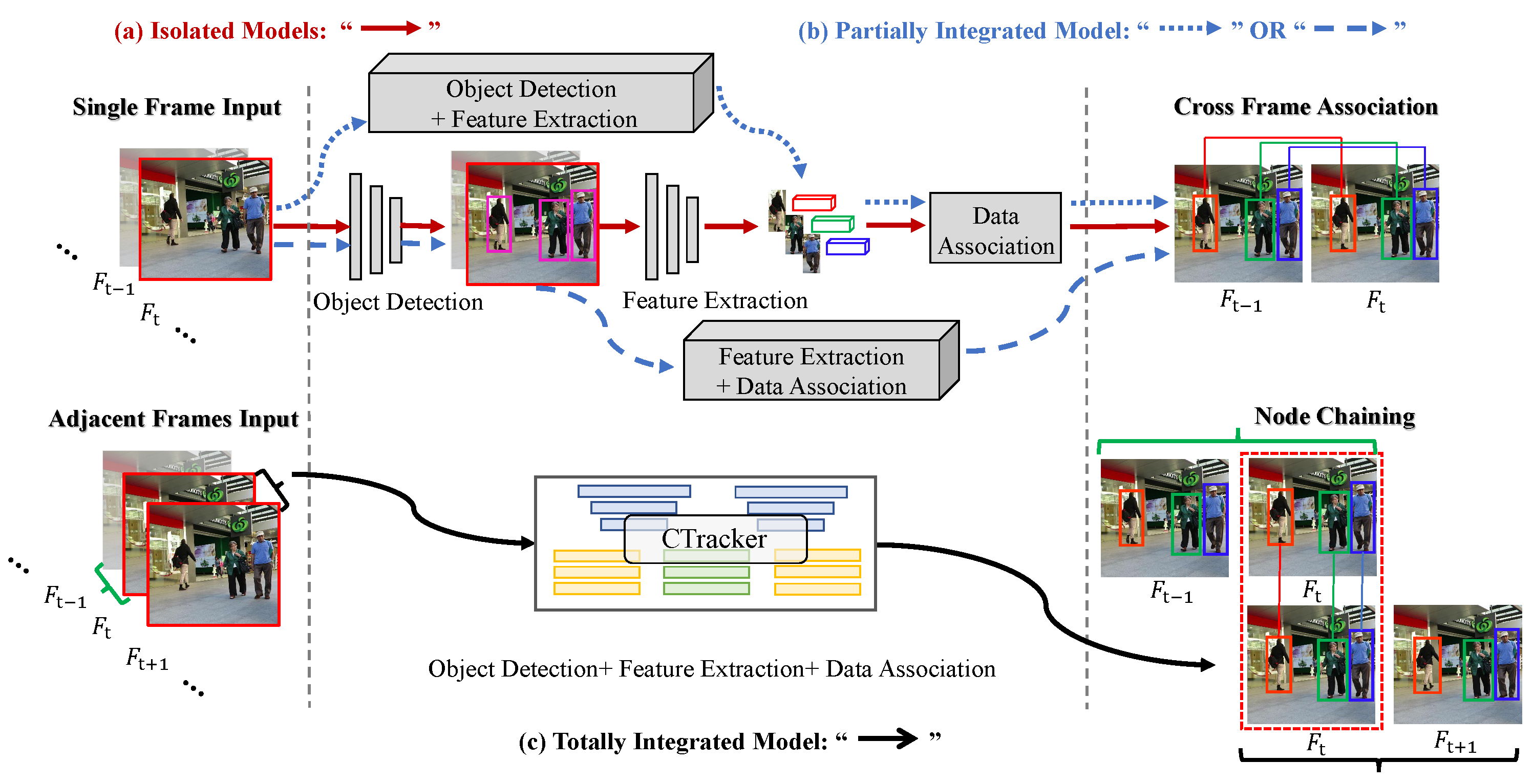}
\caption{\label{fig:comparision}\textbf{Comparison of our CTracker (Bottom) with other typical MOT methods (Top)}, which are either isolated \CRwfb{models} or partially integrated \CRwfb{models}.
Our CTracker significantly differs from other methods in two aspects: $1$) It is a totally end-to-end model using adjacent frame pair as input and generating the box pair representing the same target. $2$) 
We convert the challenging \CRwfb{cross-frame} association problem into pair-wise object detection problem.}
\label{fig:difference}
\end{figure*}

In order to solve the above problems, we propose \yang{a novel} online tracking method named \textit{Chained-Tracker} (CTracker), \wca{which unifies object detection, feature extraction and data association into a single \CRyang{end-to-end} model}. As can be seen in Fig. \ref{fig:comparision}, our novel CTracker model is \CRyang{cleaner and simpler than} the classical tracking-by-detection or partially end-to-end MOT methods. It takes adjacent frame \yang{pairs} as input to perform joint detection and tracking in a single \CRyang{regression model that simultaneously regress the paired bounding boxes for the targets that appear in both of the two adjacent frames}. 

Furthermore, we introduce a joint attention module using predicted confidence maps \CRwfb{to further improve the performance of our CTracker.  It guides the paired boxes regression branch to focus on informative spatial regions with two other branches}. One is the object classification branch, which predicts the confidence scores for \yang{the first box in the detected box pairs, and such scores are} used to guide the regression branch to focus on the foreground regions. The other one is the ID verification branch whose prediction facilitates the regression branch to focus on regions corresponding to the same target. Finally, the bounding box \wca{pairs are filtered according to the classification confidence. Then, the generated box pairs} \yang{belonging} to the adjacent frame pairs could be associated \yang{using} \CRyang{simple methods like} \wca{IoU (Intersection over Union) \yang{matching}} \cite{bochinski2017high} according to their boxes in the common frame. \pjl{In this way, the tracking process could be achieved by chaining all the adjacent frame pairs \CRyang{(\emph{i.e.} chain nodes) sequentially.}}

\pjl{\CRyang{Benefiting} from the end-to-end optimization of joint detection and tracking network, our model shows significant \CRyang{superiority} over strong competitors while \yang{remaining} simple.} 
\wca{With the temporal information of the combined \yang{features} from adjacent frames, the detector \yang{becomes} more robust, which in turn makes data association easier, and \pjl{finally} results in better tracking performance.}

The contribution of this paper can be summarized into the following aspects:

1. We propose an end-to-end online \yang{Multiple-Object} Tracking model, to optimize object detection, feature extraction and data association simultaneously. Our proposed CTracker is the first method \wfb{that} converts the challenging data association problem to \CRwfb{a} pair-wise object detection problem.

\wca{2. We design a joint attention module to highlight informative regions for box pair regression and the performance of our \yang{CTracker} is further improved.}

\wfb{3. Our online CTracker achieves state-of-the-art performance on the tracking result list with private detection of MOT16 and MOT17.}

\section{Related Work}

\subsection{Detection-based MOT Methods}

Yu \textit{et. al} \cite{yu2016poi} proposed the POI algorithm, which conducted a high-performance detector based on Faster R-CNN \cite{ren2015faster} by adding several extra pedestrian detection datasets. Chen \textit{et. al} \cite{chen2017enhancing} incorporated an \wfb{enhanced} detection model by simultaneously modeling the detection-scene relation and detection-detection relation, called EDMT. Furthermore, Henschel \textit{et. al} \cite{Henschel2017} added a head detection model to support MOT in addition to original pedestrian detection, which \wfb{also} needed extra training data and annotations. Bergmann \textit{et. al} \cite{bergmann2019tracking} proposed the Tracktor by exploiting the bounding box regression to predict the position of the pedestrian in the next frame, which was equal to modifying the detection box. However, the detection model and the tracking model in these detection-based methods are completely \textbf{independent}, which is complex and \wfb{time-consuming}. While our CTracker algorithm only needs one \textbf{integrated} model to perform detection and tracking, which is simple and efficient.

\subsection{Partially End-to-end MOT Methods}

\CRpjl{Lu \textit{et. al} \cite{lu2020retinatrack} proposed RetinaTrack, which combined detection and feature extraction in the network and used greedy bipartite matching \CRwfb{for} data association.} Sun \textit{et. al} \cite{sun2019deep} harnessed the power of deep learning for data association in tracking by jointly modeling object appearances and their affinities between different frames. Similarly, Chu \textit{et. al} \cite{chu2019famnet} designed the FAMNet to jointly optimize the feature extraction, affinity estimation and multi-dimensional assignment. \CRpjl{Li \textit{et. al} \cite{li2019tracknet} proposed TrackNet by \CRyang{using frame tubes as input to do joint detection and tracking, however the links among tubes are not modeled which limits the trajectory lengths. Moreover, the model is designed and tested only for rigid object (vehicle) tracking, leaving its generalization ability questionable.}} \wfb{\CRyang{Despite their differences}, all these methods are just \textbf{partially} end-to-end MOT methods, because they just integrated some parts of the whole model, \textit{i.e. } }\cite{lu2020retinatrack} combined the detection and feature extraction module in a network, \cite{sun2019deep,chu2019famnet} combined the feature extraction and data association module. Differently, our CTracker is a \textbf{totally} end-to-end joint detection and tracking methods, unifying the object detection, feature extraction and data association in a single model.

\subsection{Attention-assistant MOT Methods}
Chu \textit{et. al} \cite{chu2017online} introduced a Spatial-Temporal Attention Mechanism (STAM) to handle the tracking drift caused by the occlusion and interaction among targets. Similarly, Zhu \textit{et. al} \cite{zhu2018online} proposed a Dual Matching Attention Networks (DMAN) with both spatial and temporal attention mechanisms to perform the tracklet data association. Gao \textit{et. al} \cite{gao2018osmo} also \CRwfb{utilized} an attention-based appearance model to solve the inter-object occlusion. All these attention-assistant MOT methods \CRwfb{used} \wfb{a} complex attention model to optimize data association in the \textbf{local} bounding box level. While our CTracker can improve both the detection and tracking performance \CRwfb{through} the simple object-attention and identity-attention in the \textbf{global} image level, which is more efficient.

\section{Methodology}
\subsection{Problem Settings}

\wfb{Given an image \CRyang{sequence} $\{F_{t}\} _{t=1}^{N}$ with totally $N$ frames,} \yang{Multiple-Object} Tracking task aims to \wfb{\CRyang{output} all} the bounding boxes \CRyang{$\{\mathcal{G}_{t}\} _{t=1}^{N}$ and identity labels $\{\mathcal{Y}^{GT}_{t}\} _{t=1}^{N}$ for} all the objects of interest in \CRyang{all the frames where they appear.} $F_{t}\in\mathbb{R^{\mathit{c\times w\times h}}}$
indicates the $t$-th frame, \CRyang{$\mathcal{G}_{t} \subset {\mathbb{R}^4}$}
represents the \CRyang{ground-truth bounding boxes of the $K_t$ number of} targets in $t$-th
frame and \CRyang{$\mathcal{Y}^{GT}_{t} \subset \mathbb{Z}$ denotes their} identities. Most of the recent MOT algorithms divide the MOT task into three \CRyang{components}, which are \CRyang{object} detection, feature extraction and data association.
However, many researches and experiments demonstrate that the \CRyang{association's effectiveness} \wfb{relies} heavily  on the \CRwfb{performance of detection}. Therefore, in order to \CRyang{better utilize their correlation,} in this paper, we propose a novel Chained-Tracker \CRyang{(abbr. CTracker)}, which uses \CRyang{a single} network to \CRyang{simultaneously} achieve object detection, feature extraction and data association. We introduce the \CRyang{pipeline} of our CTracker in the subsection \ref{subsection:chainedtracker}. The details of the network and loss design are described separately in the subsection \ref{subsection:network} and \ref{subsection:loss}.

\begin{figure*}[t]
\centering
\includegraphics[width=0.9\columnwidth]{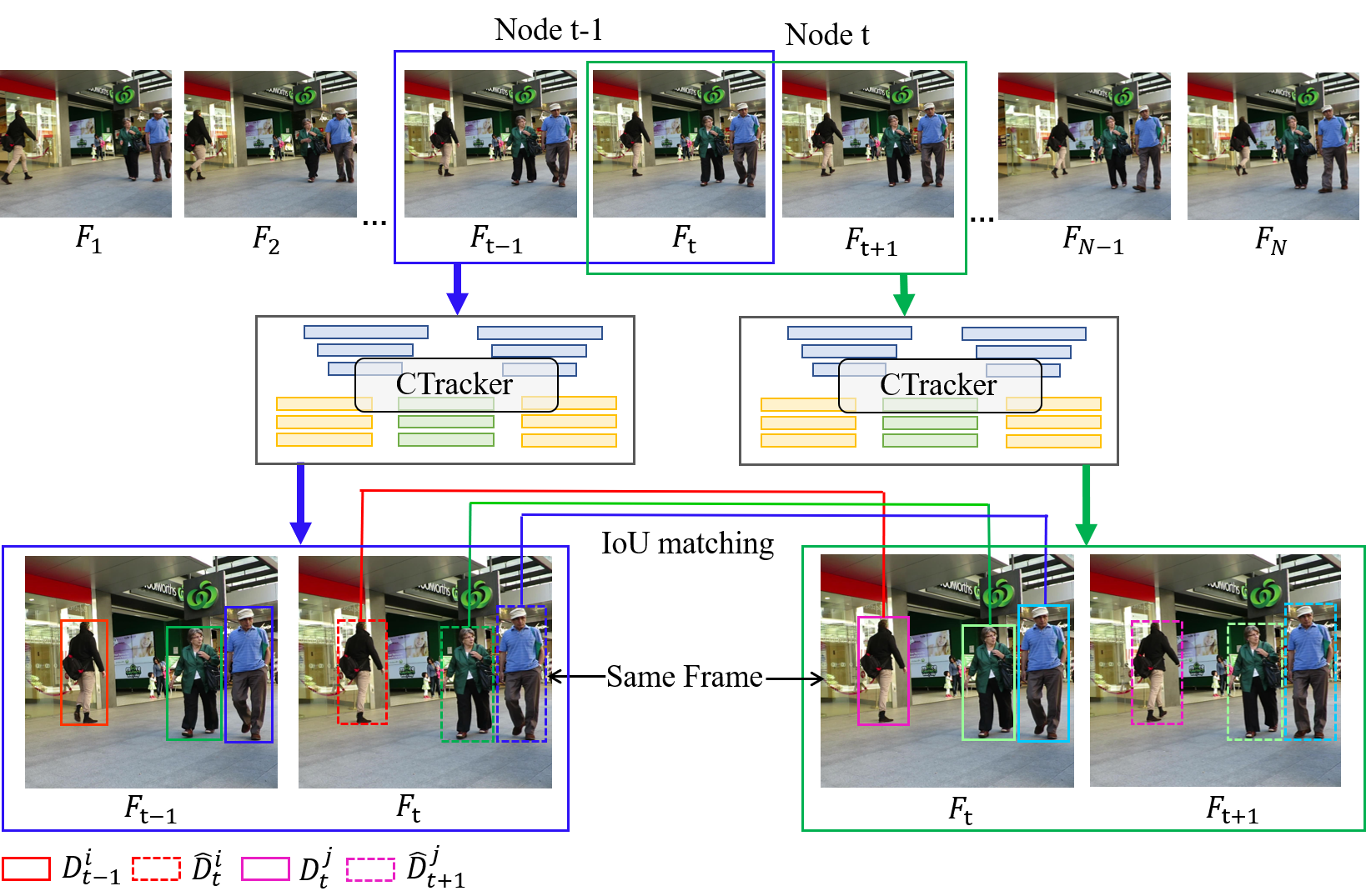}
\caption{\label{fig:model2}\textbf{Illustration \CRwfb{of} the \CRyang{node chaining}}. \CRyang{After generating bounding box pairs $ \{\mathcal{D}_{t-1},\mathcal{\hat{D}}_{t}\}$ by CTracker for two arbitrary adjacent nodes $ (F_{t-1},F_{t})$ and $ (F_{t},F_{t+1})$, we chain these two nodes by doing IoU matching on the shared common frame. Such a chaining is done sequentially over all adjacent nodes to generate long trajectories for the whole video sequence. More detailed can be found in the main text.}}
\label{fig:chain}
\end{figure*}

\subsection{Chained-Tracker Pipeline \label{subsection:chainedtracker}}

\noindent\textbf{Framework.} Different from other MOT models \CRwfb{that} only \wfb{takes} \textit{a single frame} as input, our \yang{CTracker} model requires \textit{two adjacent frames} as input, which is called a chain node. The first chain node is $ (F_1,F_2)$ and the last \CRyang{(\emph{i.e.,} the \CRwfb{$N$-th}) is} $ (F_N,F_{N+1})$. Note that $F_N$ is the last frame, \wfb{so we just take the copy version of $F_N$ as $F_{N+1}$.} \wfb{Given the node 
$(F_{t-1},F_{t})$ as input, \CRyang{CTracker} can generate \CRyang{bounding box pairs} 
$ \{(D_{t-1}^{i},\hat{D}_{t}^{i})\}_{i=1}^{\CRyang{n}_{t-1}}$ of the same \CRyang{targets appearing in both frames},}
\pjl{where $\CRyang{n}_{t-1}$ is the total pair number\CRyang{, $D_{t-1}^{i} \in \mathcal{D}_{t-1} \subset {\mathbb{R}^4}$ and $\hat{D}_{t}^{i} \in \mathcal{D}_t \subset {\mathbb{R}^4}$ denote} the two bounding boxes of the same target.
Similarly, we can also get the box pairs $ \{(D_{t}^{j},\hat{D}_{t+1}^{j})\}_{j=1}^{\CRyang{n}_{t}}$ in the next node $(F_{t},F_{t+1})$. 
As can be seen in Fig.~\ref{fig:chain}, assume that $\hat{D}_{t}^{i}$ and $ D_{t}^{j}$ \wfb{represent detected boxes of the same target located in the common frame of the adjacent nodes}, there \CRyang{shall be} only slight difference between \wfb{the} two boxes. \wfb{We can further use an extremely simple matching strategy (\CRyang{as detailed below}) to chain the two boxes, instead of using \CRyang{complicated appearance features} as in canonical MOT methods.}} \wfb{By chaining nodes \CRyang{sequentially over} the given sequence, \CRwfb{we can obtain \CRyang{long} trajectories of all the detected targets.}}

\noindent \textbf{Node chaining.} We use $ \{\mathcal{D}_{t-1},\mathcal{\hat{D}}_{t}\}$ to represent $ \{(D_{t-1}^{i},\hat{D}_{t}^{i})\}_{i=1}^{\CRyang{n}_{t-1}}$ for convenience. \CRyang{The node chaining is done as follows}. 
Firstly, in the node, \CRyang{every} detected \CRyang{bounding box $D_{1}^{i} \in \mathcal{D}_{1}$} is initialized as a tracklet with a randomly assigned identity. Secondly, for any another node $t$, we chain the adjacent nodes $(F_{t-1},F_{t})$ and $(F_{t},F_{t+1})$ by calculating the IoU \CRyang{(Intersection over Union)}  between the boxes in $\mathcal{\hat{D}}_{t}$ and $\mathcal{D}_{t}$ \CRyang{as shown in Fig. \ref{fig:chain}}, where $\mathcal{\hat{D}}_{t}$ is the last boxes set \CRyang{of} $\{\mathcal{D}_{t-1},{\mathcal{\hat{D}}_{t}}\}$ and $\mathcal{D}_{t}$ is the former boxes set \CRyang{of} $\{\mathcal{D}_{t},{\mathcal{\hat{D}}_{t+1}}\}$. Getting the IoU affinity, the detected boxes in $\mathcal{\hat{D}}_{t}$ and $\mathcal{D}_{t}$ are matched by applying the Kuhn-Munkres (KM) algorithm \cite{kuhn1955hungarian}. For each matched box \CRyang{pair} ${\hat{D}_{t}^{i}}$ and ${D_{t}^{j}}$, the tracklet \CRwfb{that} ${\hat{D}_{t}^{i}}$ belongs to is updated by appending ${D_{t}^{j}}$. Any unmatched box \CRyang{${D_{t}^{k}}$} is initialized as a new tracklet with a new identity. \CRyang{The chaining is done sequentially over all adjacent nodes and it builds long trajectories for individual targets.} 

\noindent \CRyang{\textbf{Robustness enhancement (esp. against occlusions).} To enhance the model's robustness to serious occlusions (which can make detection fail in certain frames) and short-term disappearing (followed by quick reappearing), we retain the terminated tracklets and their identities for up to $\sigma$ frames and continue finding matches for them in these frames, with the simple constant velocity prediction model \cite{wojke2017simple,peng2020tpm} for motion estimation. In greater details, suppose target $(D_{t-1}^{l},\hat{D}_{t}^{l})$ cannot find its match is node $t$, we apply the constant velocity model to predict its bounding box $P_{t+\tau}^{l}$ in frame $t+\tau$ ($1 <= \tau <= \sigma$) according to $D_{t-1}^{l}$ (not the less reliable $\hat{D}_{t}^{l}$). When we chain node $t+\tau-1$ and node $t+\tau$ with $\{\mathcal{D}_{t+\tau-1},{\mathcal{\hat{D}}_{t+\tau}}\}$ and $\{\mathcal{D}_{t+\tau},{\mathcal{\hat{D}}_{t+\tau+1}}\}$, the current set of all the predicted bounding boxes of retained targets denoted by $\mathcal{P}_{t+\tau}$, is appended to $\mathcal{\hat{D}}_{t+\tau}$ for matching with $\mathcal{D}_{t+\tau}$. If $P_{t+\tau}^{i}$ gets a match, its tracklet will be extended by linking to the new bounding boxes.}

\noindent \CRyang{\textbf{Effectiveness and limitations.} Our model is effective for handling the cases when targets appear or disappear (\emph{i.e.}, enter or leave camera view), which are quite common for MOT. When a target is not in frame $t-1$ but appears in frame $t$, it is likely that no bounding box pair for it gets generated in the chain node $(F_{t-1}, F_t)$. However, as long as this target continues to appear in frame $t+1$, it will be detected in the next chain node $(F_t, F_{t+1})$ and get a new tracklet and identity there. Similarly, if a target is in the frame $t-1$ but disappears from frame $t$, it will not be detected in node $(F_t, F_{t+1})$, resulting the termination of its tracklet in node $t-1$ or even $t-2$. Note that the chaining operation itself cannot be fully parameterized and therefore it cannot be optimized together with the regressions. Since the regression model (as detailed below) does the major work and there is no need to get feedback for it from the chaining operation, we still use the ``end-to-end'' property to describe CTracker. A pure end-to-end trainable model requires a differentiable replacement to the current IoU matching based chaining strategy.}

\subsection{Network architecture\label{subsection:network}}

\begin{figure*}[t]
\centering{}\includegraphics[width=0.95\columnwidth]{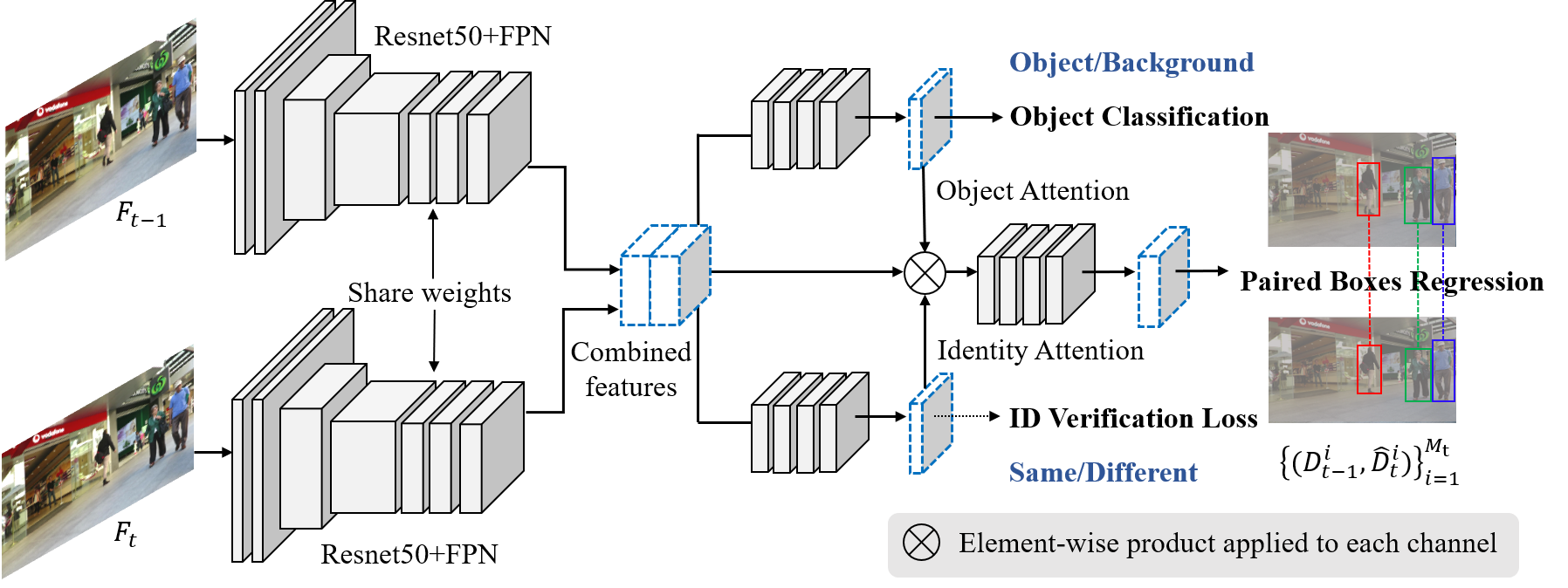}
\caption{\label{fig:model}\textbf{\CRyang{Network} architecture of \CRyang{CTracker}}. Given two adjacent frames, we firstly use \pjl{two backbone branches with tied weights} to extract the features \CRyang{for} each frame separately. Then we \CRwfb{concatenate} features of the two frames on channel level and the combined features are used to predict the paired boxes. To highlight local informative regions for paired boxes regression, \CRyang{the combined features are multiplied with the attention maps from the object classification branch and the ID verification branch.}}
\end{figure*}

\noindent \noindent\textbf{\wfb{Overview.}} Our proposed CTracker network uses two adjacent frames as input and \wfb{regresses} the bounding box pair of the same target. \CRyang{To do this, we} adopt ResNet-50 \cite{he2016deep} as \wfb{the} backbone to extract high-level semantic features. It then \wfb{integrates Feature Pyramid Networks (FPN) to generate multi-scale feature representation for subsequent prediction.} \wfb{In order to associate targets in adjacent frames}, the scale-level feature maps from \CRyang{individual frames} are firstly concatenated together, and then fed into the prediction network to \CRyang{regress bounding box pairs}. As can be seen in Fig. \ref{fig:model}, the paired boxes regression branch generates a box pair for each target, and the object classification branch predicts a score for each pair \CRwfb{indicating} the confidence of being foreground. To \CRwfb{help} the paired boxes regression branch \CRwfb{to} \CRyang{avoid the distraction by irrelevant yet confusing information, \wfb{the} object classification branch and the extra ID verification branch are used for attention guidance.}

\noindent \textbf{\wfb{Paired Boxes Regression.}} \wfb{Inspired by} \CRwca{predicting the offsets relative to pre-defined (default) anchor boxes in} object detection, we propose Chained-Anchors for the paired boxes regression branch to regress two boxes simultaneously. \CRwca{As a novel natural derivative of the anchors used in most object detection methods, Chained-Anchors are densely arranged on a spatial grid, each of them allows predicting two bounding boxes of the same object instance in two adjacent frames.} In order to handle the large scale variation in real scenes, the K-means clustering \CRyang{as used in} \cite{yolov2} is conducted on all ground-truth bounding boxes in the dataset \CRyang{for getting the scales of chained-anchors.} And each \CRyang{cluster} is assigned to the corresponding level of FPN for later scale specific predictions. The detected bounding box pairs are firstly post-processed with \CRyang{soft-NMS} \cite{softnms} according to the IoU of the first box in each pair, and then filtered based on the confidence scores from the classification branch. Finally, the remaining box pairs are chained into the whole tracking \CRyang{trajectories} using \wfb{the} method described in Sec.~\ref{subsection:chainedtracker}. To keep our model simple, both the paired boxes regression branch and the classification branch only stack four consecutive 3$\times$3 \CRwfb{Conv} layers interleaved with ReLU activations before the \CRwfb{final} convolution \CRwfb{layer}.

\noindent \textbf{Joint Attention Module.} 
\CRyang{We} design an attention mechanism based component called Joint Attention Module (JAM) to highlight local informative regions in the combined features before the regression branch. \CRyang{As shown from the right of Fig. \ref{fig:model},} the ID verification branch \CRyang{is introduced} to \CRwfb{get} confidence scores, indicating whether the two boxes in the detected pair belong to the same target. Then both the predicted confidence map of ID verification branch and object classification branch are used as attention \CRyang{maps}. Note that the guidance from the two branches is complementary, the confidence maps from the classification branch focuses on foreground regions while the prediction from the ID verification branch is used to highlight the features of the same target. 

\begin{figure}[t!]
\begin{centering}
\includegraphics[width=0.6\textwidth]{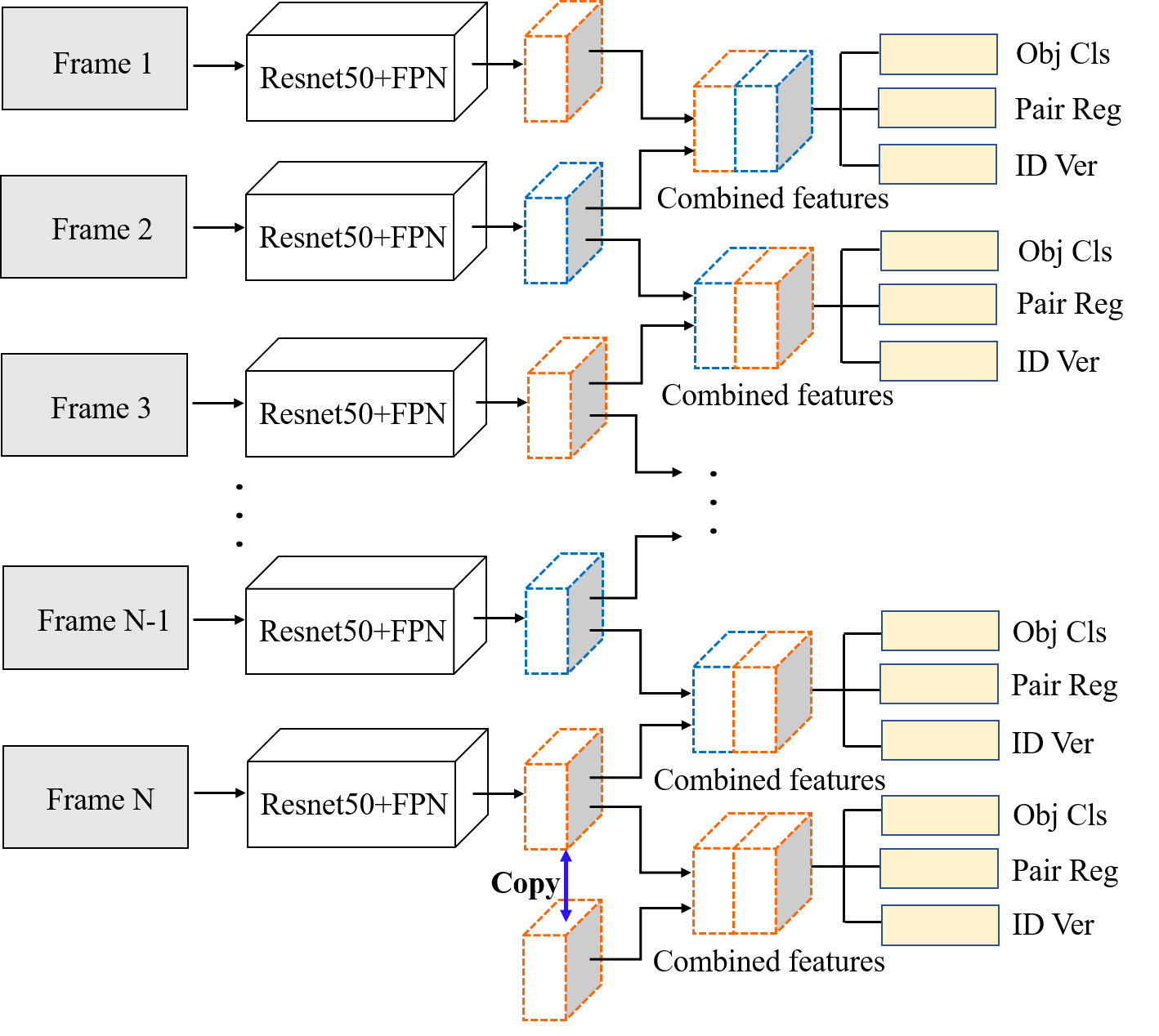} 
\par\end{centering}
\caption {\textbf{Memory sharing mechanism in our CTracker}. The extracted features of each frame (except the first one) are firstly used in the current chain node, and then can be saved and reused in the next chain node. Note that when making inference for the last node, the features of the last frame $N$ is also reused as the features of the hypothetical frame $N+1$ to avoid the \CRwfb{repeated} computation for frame $N$. \label{fig:memory}}
\end{figure}

\noindent \textbf{Feature Reuse.} Since the input of the network contains two adjacent frames, the common frame of \CRyang{two} adjacent nodes \CRyang{has to be used twice}
\pjl{in the tracking process.}
\wfb{To avoid the nearly double cost of computation and memory in inference, we propose a Memory Sharing Mechanism (MSM) to temporarily save the extracted features of the current frame and reuse them until the next node is processed, as shown in Fig. \ref{fig:memory}.} Besides, in order to make inference for the last node, we make a copy of frame $N$ as the hypothetical frame $N+1$. To further avoid the \CRwfb{repeated} computation for \CRwfb{the} frame $N+1$, we also apply the trick of feature resue to frame $N$, and the feature of frame $N$ is copied as the feature of the hypothetical frame $N+1$. We demonstrate that the proposed MSM can reduce \CRwfb{almost half of the overall computation and time cost.}

\subsection{Label \CRyang{Assignment} and Loss Design\label{subsection:loss}}

\CRyang{For an arbitrary chain node ($F_t$, $F_{t+1}$), let $A_t^i=(x^{t,i}_a, y^{t,i}_a, w^{t,i}_a, h^{t,i}_a)$ denote its $i$-th chained-anchor (where $x^{t,i}_a$ and $y^{t,i}_a$ are the box center coordinates; $w^{t,i}_a$ and $h^{t,i}_a$ are the width and height, respectively), we adopt a ground-truth bounding box matching strategy similar to that of SSD \cite{liu2016ssd}. We use a matrix $M$ to denote the result of such a matching. If $G_t^j$ is the corresponding ground-truth bounding box in $F_t$ for $A_t^i$, which is judged by the IoU ratio (higher than a threshold $T_p$), then we have $M_{ij} = 1$. If the IoU ratio is lower than another smaller threshold $T_n$, then $M_{ij} = 0$. Based on $M$, we can assign the ground-truth label $c_\mathrm{cls}^i$ to CTracker's classification branch for $A_t^i$ as:} \begin{equation}
\CRyang{c}_\mathrm{cls}^i = \left\{\begin{matrix}
1,\:  & \mathrm{if \: \Sigma}_{j=1}^{\CRyang{K_t}} \CRyang{M_{ij}} = 1, \\ 
0,\: & \mathrm{if \: \Sigma}_{j=1}^{\CRyang{K_t}} \CRyang{M_{ij}} = 0,
\end{matrix}\right.
\end{equation}  
where \CRyang{$K_t$} is the total number of ground-truth \CRyang{bounding boxes for} frame $F_t$.

\CRyang{With $A_t^i$, suppose the predicted pair of bounding boxes are $(D_t^i, \hat{D}_{t+1}^i)$ and the corresponding ground-truth bounding boxes are $(G_t^j, G_{t+1}^k)$ when they exist, the ID verification branch of CTracker shall get its ground-truth label as:}
\begin{equation}
\CRyang{c}_\mathrm{id}^i = \left\{\begin{matrix}
1,\:  & \mathrm{if} \: \CRyang{c}_\mathrm{cls}^i = 1 \;\mathrm{\mathbf{ and }} \; \mathcal{I}[G_t^j] = \mathcal{I}[G_{t+1}^k], \\ 
0,\: & \mathrm{otherwise},
\end{matrix}\right.
\end{equation}
\CRyang{where $\mathcal{I}[\cdot]$ represents the identity of the target in the bounding box.}

\CRyang{We follow Faster R-CNN \cite{faster} to regress offsets of $(D_t^i, \hat{D}_{t+1}^i)$ w.r.t. $A_t^i$, where $D_t^i= (x^{t,i}_d, y^{t,i}_d, w^{t,i}_d, h^{t,i}_d)$. Let $(\Delta^{t,i}_d, \Delta^{t+1,i}_{\hat{d}})$ denote these offsets and $(\Delta^{t,j}_g, \Delta^{t+1,k}_g)$ be the offsets for the ground-truths, we list the details of $\Delta^{t,i}_d = (\Delta^{t,i}_{d,x}, \Delta^{t,i}_{d,y}, \Delta^{t,i}_{d,w},$ $\Delta^{t,i}_{d,h})$ as an example (the others are similar):}
\begin{equation}
\begin{aligned}
    \CRyang{\Delta^{t,i}_{d,x} = (x^{t,i}_d - x^{t,i}_a)/w^{t,i}_a,} & \quad \CRyang{\Delta^{t,i}_{d,y} = (y^{t,i}_d - y^{t,i}_a)/h^{t,i}_a, } \\ \CRyang{\Delta^{t,i}_{d,w} = \mathrm{log}(w^{t,i}_d/w^{t,i}_a),} & \quad \CRyang{\Delta^{t,i}_{d,h} = \mathrm{log}(h^{t,i}_d/h^{t,i}_a).}
\end{aligned}
\end{equation}
\CRyang{The loss for the paired boxes regression branch is defined as follows:}
\begin{equation}
\begin{split}
\label{eq:l1}
&\CRyang{L_{reg}(\Delta^{t,i}_d, \Delta^{t+1,i}_{\hat{d}}, \Delta^{t,j}_g, \Delta^{t+1,k}_g)} \\
& \CRyang{=  \sum_{\CRyang{l}\in\{x,y,w,h\}}\left[{\mathrm{smooth}_{L_1}}(\Delta^{t,i}_{d,l} - \Delta^{t,j}_{g,l}) + {\mathrm{smooth}_{L_1}}(\Delta^{t+1,i}_{\hat{d},l} - \Delta^{t+1,k}_{g,l})\right]/8,}
\end{split}
\end{equation}
\CRyang{where ${\mathrm{smooth}_{L_1}}$ is the smooth $L_{1}$ loss. }

\CRyang{The total loss of CTracker is}
\begin{equation}
\CRyang{L_{all} = \sum_{t,i} \left[ L_{reg}(\Delta^{t,i}_d, \Delta^{t+1,i}_{\hat{d}}, \Delta^{t,j}_g, \Delta^{t+1,k}_g) + \alpha \mathcal{F}(p^i_{cls}, c^i_{cls}) +\beta \mathcal{F}(p^i_{id}, c^i_{id})\right],}
\label{eq:totalloss}
\end{equation}
\CRyang{where $\mathcal{F}(p^i_{cls}, c^i_{cls})$ and $\mathcal{F}(p^i_{id}, c^i_{id})$ are the focal losses \cite{lin2017focal} for the classification branch and the ID verification branch (for mitigating the sample imbalance problem), respectively, with $p^i_{cls}$ and $p^i_{id}$ denoting their predictions (confidence scores); $\alpha$ and $\beta$ are the weighting factors.}

\section{Experiment}
\subsection{Datasets and Evaluation Metrics}
We conduct the experiments on two public datasets: MOT16 \cite{milan2016mot16} and MOT17. which contain \wfb{the} same image sequences including 7 training sequences and 7 test sequences. \CRpjl{However, MOT16 and MOT17 contain different detection input, and different ground-truth labels (bounding boxes and identities), which would influence the training of CTracker.} \wfb{In} public detection, MOT16 includes DPM \cite{felzenszwalb2010object} detector while MOT17 includes DPM, \CRyang{Faster R-CNN} \cite{ren2015faster} and SDP \cite{yang2016exploit} detectors. For a fair comparison with other methods, we trained two models separately using the training data from MOT16 and MOT17, and separately applied the two models on the MOT16 test set and MOT17 test set.

In the MOTChallenge benchmark, tracking performance is measured by the widely used CLEAR MOT Metrics \cite{bernardin2008evaluating}, including \yang{Multiple-Object} Tracking Accuracy (MOTA), \yang{Multiple-Object} Tracking Precision (MOTP), the total number of False Negatives (FN), False Positives (FP), Identity Switches (IDS), and the percentage of Mostly Tracked Trajectories (MT), Mostly Lost Trajectories (ML). ID F1 Score (IDF1) is also used to measure the trajectory identity accuracy. Among these metrics, MOTA is the primary metric to measure the overall detection and tracking performance. In addition, we use Tracker Speed in Frames Per Seconds (Hz) to measure the tracking speed of all methods. 

\subsection{Implementation Details}
All the experiments are implemented on the PyTorch framework. During training, the ground-truth boxes with a visible score above 0.1 are selected to train the network. In order to avoid overfitting, we use several data augmentation strategies such as photometric distortions, random flip and random crop. The same augmentation operation is guaranteed to apply for each image in the same training pair. Then the augmented image pair \CRyang{are} resized or padded to the half of their original images' shorter side. \CRyang{We also add a novel data augmentation strategy in the temporal dimension to form chain nodes: instead of always choosing two adjacent frames, we sample two frames close to each other with a random temporal gap (1 to 3 frames).} 

As a speed-accuracy trade-off, we use the Resnet50 \cite{he2016deep} network as the backbone in all the following experiments. All trainable weights except the BN parameters in Resnet50 are trained end-to-end using \CRwfb{the} Adam optimizer. We initialize the parameters for all the newly added convolutional layers with the Kaiming initialization method in \cite{he2015delving} and set the initial learning rate to $5\times e^{-5}$. The model training process takes 100 epochs with the batch size of 8 (4 training pairs). \CRyang{The weighting factors $\alpha$ and $\beta$ in the loss function are both set to 1. }In the anchor matching stage, we use 0.5 for the positive threshold and 0.4 for the negative threshold. For paired boxes post-processing, we use a threshold of 0.7 for the soft-nms, and then further filter remaining pairs with the confidence threshold of 0.4. In the chaining stage, the IoU matching threshold is 0.5, and the retention threshold of $\sigma$ is 10.

\begin{table}[t!]
\renewcommand\arraystretch{1.2}
\centering

\caption{\textbf{Ablation study on MOT17 test dataset}.}\label{tab:tab1}
\scriptsize{
\setlength{\tabcolsep}{0.8mm}{
\begin{tabular}{|c|cccccccc|}

\hline
Method & MOTA$\uparrow$ & IDF1$\uparrow$ & MOTP$\uparrow$ & MT$\uparrow$ & ML$\downarrow$ & FP$\downarrow$ & FN$\downarrow$ & IDS$\downarrow$\\
\hline
Baseline& 64.4 & 51.6 & 78.2 & 28.5\% & 28.0\% & {\bf 16089} & 178704 & 6336\\
Baseline+ObjAtten& 66.0 & 55.7 & {\bf 78.8} & 31.3\% & 24.5\% & 17724 & 168522 & 5595\\
Baseline+ObjAtten+IDVer& 65.6 & 55.2 & 78.3 & {\bf 32.6\%} & 24.7\% & 25815 & 162489 & 5769\\
{\bf Baseline+JointAtten}& {\bf 66.6} & {\bf 57.4} & 78.2 & 32.2\% & {\bf 24.2\%} & 22284 & {\bf 160491} & {\bf 5529}\\
\hline
\end{tabular}}}
\end{table}

\subsection{Ablation Study}

\noindent\textbf{Performance analysis.} \wca{We \CRyang{compare} the following \CRyang{models on MOT17 dataset to show} the effectiveness of \CRyang{CTracker's} parts:}

\noindent\wca{(1) \emph{Baseline}. \CRyang{It only covers the classification branch} and the paired boxes regression branch, without guidance from any attention map. This is the simplest implementation of our CTracker.}

\noindent\wca{(2) \emph{Baseline+ObjAtten}. \CRyang{In addition to the Baseline, the} predicted confidence map of the object classification branch is used as an attention map, which is multiplied to the combined features before the paired boxes regression branch.}

\noindent\CRpjl{(3) \emph{Baseline+ObjAtten+IDVer}. Except for the object classification branch with attention map and the paired boxes regression branch, we add the ID verification branch but do not use it as attention guidance.}

\noindent\wca{(4) \emph{Baseline+JointAtten (CTracker)}. This is the \CRyang{full} version of our approach.} 

\CRyang{Results presented in Table~\ref{tab:tab1} show} that:

(1)~\emph{Baseline+ObjAtten} performs significantly better than \emph{Baseline}, which proves the effectiveness of the object attention operation. By applying the object classification branch as the attention map of the paired boxes regression branch, \wfb{we can get more accurate bounding boxes. There is a significant improvement of MOTA, which increases from 64.4 to 66.0 and MOTP also increases from 78.2 to 78.8. The more accurate bounding boxes also result in better performance of data association, with IDF1 increasing from 51.6 to 55.7.}

\CRpjl{(2)~\emph{Baseline+ObjAtten+IDVer} performs slightly worse than \emph{Baseline+ObjAtten}. Simply adding the independent ID verification branch is weak due to the lack of bounding boxes information. Reliable identification needs good bounding boxes.}

\begin{figure*}[t!]
\centering{}\includegraphics[width=0.98\columnwidth]{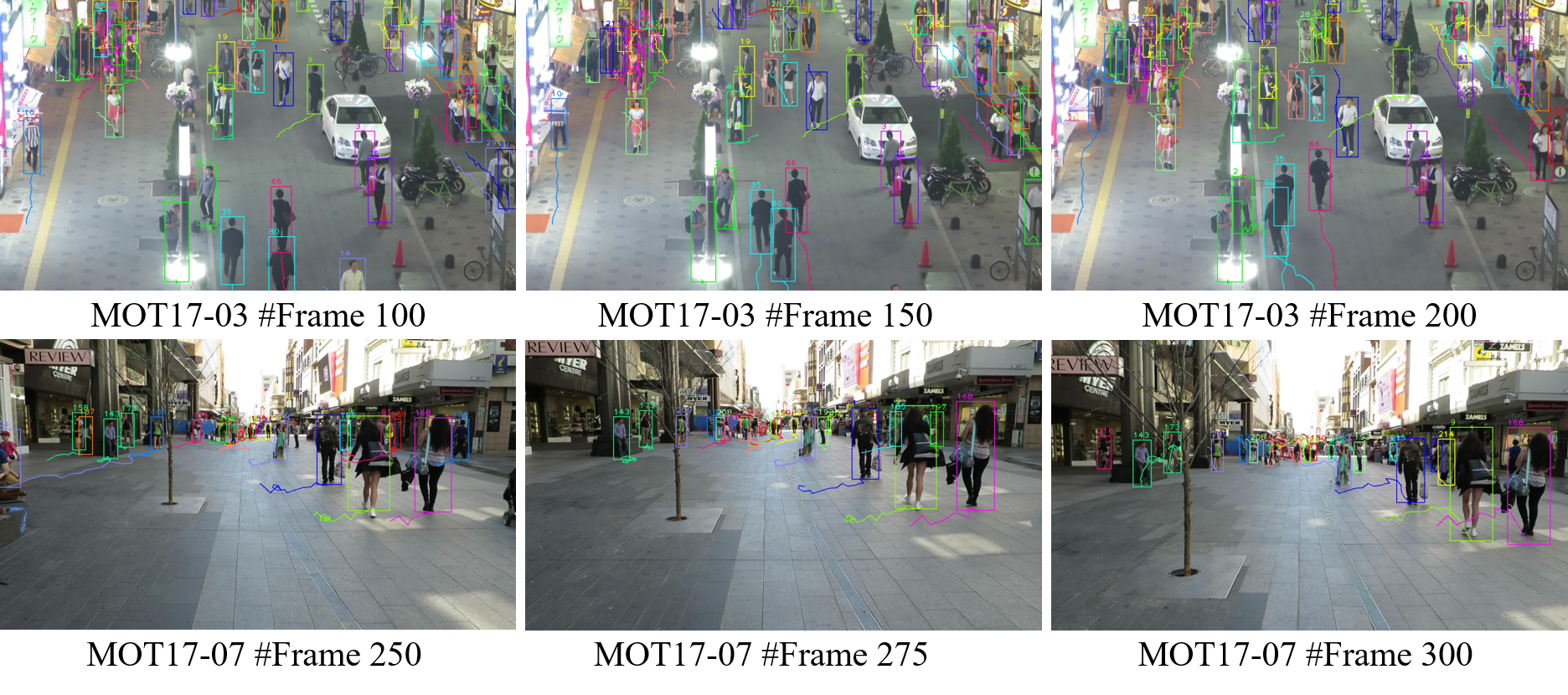}
\caption{\label{fig:result}\textbf{Qualitative results of our \yang{CTracker} on MOT17 test dataset}. MOT17-03 sequence is captured by a static camera and MOT17-07 sequence is captured by a moving camera. The \CRwfb{detected} bounding boxes and the tracking trajectory with the same identity are displayed by the same color.}
\end{figure*}

(3)~\emph{Baseline+JointAtten} further outperforms \emph{Baseline+ObjAtten}, indicating that the ID attention operation is also \CRyang{beneficial}. By adding the ID verification branch and using it as another guidance of the paired boxes regression branch, the association of the regressed bounding boxes is more accurate. Though MOTA is only improved by 0.6, the IDF1 is improved by 1.7, and IDF1 can better reflect the accuracy of data association more clearly. On the other hand, by adding the ID attention, the model pays more attention to the data association and sacrifices slightly of the regression bounding box precision, thus the MOTP is decreased from 78.8 to 78.2. Qualitative results of CTracker are illustrated in Fig.~\ref{fig:result}.

\begin{table}[t!]
\renewcommand\arraystretch{1.2}
\centering
\caption{\textbf{Time cost analysis of CTracker}.}\label{tab:time}

\scriptsize{
\setlength{\tabcolsep}{0.8mm}{
\begin{tabular}{|c|ccc|c|}
\hline
\multirow{2}{*}{Methods}&\multicolumn{4}{c|}{Time cost (ms)}\\ \cline{2-5}
& Backbone & Prediction & Chaining& Total\\ \hline
CTracker-Det&  80.27 & 38.78 & - & 119.05 \\
CTracker w/o MSM & 154.53 & 66.93 & 2.10  & 223.56\\
{\bf CTracker}&  80.29 & 65.71& 2.10 & 148.10\\
\hline
\end{tabular}}}
\end{table}

\noindent\textbf{Time cost analysis.} We \CRwfb{analyze} the inference speed for each module in CTracker, displayed in Table~\ref{tab:time}. The time cost is measured for 1080$\times$1920 images using single Tesla P40 and cuDNN v7 with Intel Xeon E5-2699v4@2.20GHz. In Table~\ref{tab:time}, CTracker-Det only predicts boxes for a single frame, which is the initial detection network of CTracker. Since nearly 70\% of the forward time is spent on the backbone network, our original CTracker costs about \CRwfb{double-time} to perform joint detection and tracking compared with the initial detection network, the time increasing from 119.05 ms to 223.56 ms. With the help of the proposed Memory Sharing Mechanism (MSM) in Sec.~\ref{subsection:network}, we achieve a faster joint detection and tracking model with only 29.05 ms extra cost compared with the detection network. There is just a small \CRwfb{increase} of time from 119.05 ms to 148.10 ms. To some extent, 29.05 ms per frame means the tracking module \CRyang{runs at} 34.4 \CRyang{FPS}, demonstrating the efficiency of our online approach.

\subsection{Benchmark Evaluation}
We compare our \yang{CTracker} approach with other MOT methods on both MOT16 and MOT17 test datasets. For comparison, we trained our model separately using the MOT16 training data and MOT17 training data. Table~\ref{tab:mot16} and Table~\ref{tab:mot17} compare the tracking results of all the methods separately on MOT16 and MOT17 test dataset. From Table~\ref{tab:mot16} and Table~\ref{tab:mot17} we can find that:

\begin{table}[t!]
\renewcommand\arraystretch{1.2}
\centering
\caption{\textbf{Comparisons of tracking results on MOT16 test dataset}.}\label{tab:mot16}
\scriptsize{
\setlength{\tabcolsep}{0.8mm}{
\begin{tabular}{|c|c|cccccccc|c|}

\hline
\multicolumn{11}{|c|}{Public Detection} \\
\hline
Process & Method & MOTA$\uparrow$ & IDF1$\uparrow$ & MOTP$\uparrow$ & MT$\uparrow$ & ML$\downarrow$ & FP$\downarrow$ & FN$\downarrow$ & IDS$\downarrow$ & Hz$\uparrow$\\
\hline
\multirow{4}{*}{Offline}& MHT-bLSTM~\cite{kim2018multi} & 42.1 & 47.8 & 75.9 & 14.9\% & 44.4\% & 11637 & 93172 & 753 & {\bf 1.8}\\
& Quad-CNN~\cite{son2017multi} & 44.1 & 38.3 & 76.4 & 14.6\% & 44.9\% & {\bf 6388} & 94775 & 745 & {\bf 1.8}\\
& EDMT~\cite{chen2017enhancing} & 45.3 & 47.9 & 75.9 & 17.0\% & {\bf 39.9\%} & 11122 & 87890 & 639 & {\bf 1.8}\\
& LMP~\cite{tang2017multiple} & {\bf 48.8} & {\bf 51.3}  & {\bf 79.0} & {\bf 18.2\%} & 40.1\% & 6654 & {\bf 86245} & {\bf 481} & 0.5\\
\hline
\multirow{5}{*}{Online}& CDA-DDAL~\cite{bae2018confidence} & 43.9 & 45.1 & 74.7 & 10.7\% & 44.4\% & 6450 & 95175 & 676 & -\\
& STAM~\cite{chu2017online} & 46.0 & 50.0 & 74.9 & 14.6\% & 43.6\% & 6895 & 91117 & {\bf 473} & -\\
& DMAN~\cite{zhu2018online} & 46.1 & {\bf 54.8} & 73.8 & 17.4\% & 42.7\% & 7909 & 89874 & 532 & -\\
& MOTDT~\cite{chen2018real} & 47.6 & 50.9 & 74.8 & 15.2\% & 38.3\% & 9253 & 85431 & 792 & {\bf 20.6}\\
& Tracktor~\cite{bergmann2019tracking} & {\bf 54.4} & 52.5 & {\bf 78.2} & {\bf 19.0\%} & {\bf 36.9\%} & {\bf 3280} & {\bf 79149} & 682 & -\\
\hline
\hline
\multicolumn{11}{|c|}{Private Detection} \\
\hline
Process & Method & MOTA$\uparrow$ & IDF1$\uparrow$ & MOTP$\uparrow$ & MT$\uparrow$ & ML$\downarrow$ & FP$\downarrow$ & FN$\downarrow$ & IDS$\downarrow$ & Hz$\uparrow$\\
\hline
\multirow{3}{*}{Offline}& NOMT~\cite{choi2015near} & 62.2 & {\bf 62.6} & {\bf 79.6} & 32.5\% & 31.1\% & {\bf 5119} & 63352 & {\bf 406} & 11.5\\
& MCMOT-HDM~\cite{lee2016multi} & 62.4 & 51.6 & 78.3 & 31.5\% & 24.2\% & 9855 & 57257 & 1394 & {\bf 34.9}\\
& KDNT~\cite{yu2016poi} & {\bf 68.2} & 60.0 & 79.4 & {\bf 41.0\%} & {\bf 19.0\%} & 11479 & {\bf 45605} & 933 & 0.7\\
\hline
\multirow{5}{*}{Online}& EAMTT~\cite{sanchez2016online} & 52.5 & 53.3 & 78.8 & 19.0\% & 34.9\% & {\bf 4407} & 81223 & 910 & 12.0\\
& DeepSORT~\cite{wojke2017simple} & 61.4 & 62.2 & 79.1 & 32.8\% & {\bf 18.2\%} & 12852 & 56668 & {\bf 781} & 20.0\\
& CNNMTT~\cite{mahmoudi2019multi} & 65.2 & 62.2 & 78.4 & 32.4\% & 21.3\% & 6578 & 55896 & 946 & 11.2\\
& POI~\cite{yu2016poi} & 66.1 & {\bf 65.1} & {\bf 79.5} & {\bf 34.0\%} & 20.8\% & 5061 & 55914 & 805 & 9.9\\
& {\bf \yang{CTracker} (Ours)} & {\bf 67.6} & 57.2 & 78.4 & 32.9\%& 23.1\% & {8934} & {\bf 48305} & 1897 & {\bf 34.4}\\
\hline
\end{tabular}}}
\end{table}

\begin{table}[t!]
\renewcommand\arraystretch{1.2}
\centering
\caption{\textbf{Comparisons of tracking results on MOT17 test dataset}.}\label{tab:mot17}
\scriptsize{
\setlength{\tabcolsep}{0.55mm}{
\begin{tabular}{|c|c|cccccccc|c|}

\hline
\multicolumn{11}{|c|}{Public Detection} \\
\hline
Process & Method & MOTA$\uparrow$ & IDF1$\uparrow$ & MOTP$\uparrow$ & MT$\uparrow$ & ML$\downarrow$ & FP$\downarrow$ & FN$\downarrow$ & IDS$\downarrow$ & Hz$\uparrow$\\
\hline
\multirow{4}{*}{Offline} & MHT-bLSTM~\cite{kim2018multi} & 47.5 & 51.9 & {\bf 77.5} & 18.2\% & 41.7\% & 25981 & 268042 & 2069 & {\bf 1.8}\\
& EDMT~\cite{chen2017enhancing} & 50.0 & 51.3 & 77.3 & {\bf 21.6\%} & 36.3\% & 32279 & {\bf 247297} & 2264 & {\bf 1.8}\\
& JCC~\cite{keuper2018motion} & 51.2 & {\bf 54.5} & 75.9 & 20.9\% & 37.0\% & 25937 & 247822 & {\bf 1802} & -\\
& FWT~\cite{Henschel2017} & {\bf 51.3} & 47.6 & 77.0 & 21.4\% & {\bf 35.2\%} & {\bf 24101} & 247921 & 2648 & -\\
\hline
\multirow{3}{*}{Online}& DMAN~\cite{zhu2018online} & 48.2 & {\bf 55.7} & 75.9 & 19.3\% & 38.3\% & 26218 & 263608 & 2194 & -\\
& MOTDT~\cite{chen2018real} & 50.9 & 52.7 & 76.6 & 17.5\% & {\bf 35.7\%} & 24069 & 250768 & 2474 & {\bf 20.6}\\
& Tracktor~\cite{bergmann2019tracking} & {\bf 53.5} & 52.3 & {\bf 78.0} & {\bf 19.5\%} & 36.6\% & {\bf 12201} & {\bf 248047} & {\bf 2072} & -\\
\hline
\hline
\multicolumn{11}{|c|}{Private Detection} \\
\hline
Process & Method & MOTA$\uparrow$ & IDF1$\uparrow$ & MOTP$\uparrow$ & MT$\uparrow$ & ML$\downarrow$ & FP$\downarrow$ & FN$\downarrow$ & IDS$\downarrow$& Hz$\uparrow$\\

\hline
\multirow{3}{*}{Online}& Tracktor+CTdet ~\cite{bergmann2019tracking} & 54.4 & 56.1 & 78.1 & 25.7\% & 29.8\% & 44109 & 210774 & 2574&-\\
& DeepSORT~\cite{wojke2017simple} & 60.3 & {\bf 61.2} & {\bf 79.1} & 31.5\% & {\bf 20.3\%} & 36111 & 185301 & {\bf 2442}&20.0\\
& {\bf \yang{CTracker} (Ours)} & {\bf 66.6} & 57.4 & 78.2 & {\bf 32.2\%}& 24.2\% & {\bf 22284} & {\bf 160491} & 5529&{\bf 34.4}\\
\hline
\end{tabular}}}
\end{table}

(1)~In the private detection part of both MOT16 and MOT17, our CTracker significantly outperforms existing online MOT methods in terms of MOTA. In MOT16, the MOTA of our approach is only 0.6 lower than the best offline method KDNT~\cite{yu2016poi}, while it is 1.5 higher than its online version POI~\cite{yu2016poi}. In addition, KDNT and POI use many extra training data, including ETHZ pedestrian dataset~\cite{ess2008mobile}, Caltech pedestrian dataset~\cite{dollar2009pedestrian} and their own collected surveillance dataset~\cite{yu2016poi}. While we only use the training data of MOT16. MOTA is the primary metric reflecting the overall detection and tracking performance, which proves the effectiveness of our approach. 

(2)~In the public detection part, Tracktor~\cite{bergmann2019tracking} performs the best in terms of MOTA. To have a comparison with Tracktor using the same detection result, we reproduce Tracktor using its code. Tracktor+CTdet in Table~\ref{tab:mot17} is the tracking result of Tracktor using the detection result of our CTracker. Compared with the results of public detection, the MOTA of Tracktor+CTdet increases from 53.5 to 54.4 and IDF1 increases from 52.3 to 56.1, which indicates that the performance of our detection is better than the public detection. Besides, our CTracker outperforms Tracktor+CTdet in terms of all the metrics \CRwfb{except} IDS, which further proves the superior tracking performance of our CTracker. 

(3)~On the other hand, to keep the simplicity and efficiency of our CTracker, we abandon using the patch-level ReID features of the detected boxes like other MOT methods to enhance cross-frame data association. Thus, the IDF1 and IDS of our CTracker approach are lower than several methods. \CRpjl{We conduct an extra experiment by adding features, introduced in the supplementary.} To further prove the efficiency of our approach, we compare the time cost of CTracker with other state-of-the-art MOT methods on the MOT16 and MOT17 benchmark, as shown in the Hz column of Tabel~\ref{tab:mot16} and Tabel~\ref{tab:mot17}. From Tabel~\ref{tab:mot16} and Tabel~\ref{tab:mot17} we can find that CTracker achieves the best tracking speed among all online MOT methods, although the fastest offline method runs at a similar tracking speed as our CTracker, but has a much lower MOTA than our CTracker, demonstrating the effectiveness and efficiency of our approach.

\section{Conclusion\label{section:conclusion}}

We designed a novel joint multiple-object detection and tracking framework named Chained-Tracker in this paper, which is the first totally end-to-end solution as far as we are aware. Different from existing methods, we use two adjacent frames as the input of our network, which is called a chain node. The network regresses a pair of bounding boxes for the same target in the two adjacent frames, guided by a simple yet novel joint attention module: an interplay of detection-driven object attention and ID verification-injected identity attention. Using the simple IoU information, two adjacent and overlapping nodes can be chained by their boxes in the common frame. The tracking trajectories can be generated by alternately applying the paired boxes regression and node chaining. Extensive experiments on widely used MOT benchmarks demonstrate the superiority of our approach in terms of both effectiveness and efficiency.

\bibliographystyle{splncs}
\bibliography{egbib}


\pagestyle{headings}
\mainmatter
\def\ECCVSubNumber{100}  

\title{Chained-Tracker: Chaining Paired Attentive Regression Results for End-to-End Joint Multiple-Object Detection and Tracking (Supplementary Material)} %

\titlerunning{Supplementary Material for ``Chained-Tracker''}
%
\author{Jinlong Peng\inst{1} $^\star$ \and
Changan Wang\inst{1}
\thanks{\textit{Equal contribution.}}
\and
Fangbin Wan\inst{2} \and
Yang Wu\inst{3} 
\thanks{\textit{Corresponding author: Yang Wu (wuyang0321@gmail.com)}}
\and
Yabiao Wang\inst{1} \and
Ying Tai\inst{1} \and
Chengjie Wang\inst{1} \and
Jilin Li\inst{1} \and
Feiyue Huang\inst{1} \and
Yanwei Fu\inst{2}
}

\authorrunning{J. Peng et al.}
%
\institute{Tencent Youtu Lab \quad
\email{\{jeromepeng, changanwang, caseywang, yingtai, jasoncjwang, jerolinli, garyhuang\}@tencent.com}
\and Fudan University \quad
\email{\{fbwan18, yanweifu\}@fudan.edu.cn}
\and
Nara Institute of Science and Technology \quad
\email{yangwu@rsc.naist.jp}
}
\maketitle

\section{Overview}

This supplementary material includes:

\noindent(1) The detailed design of the CTracker network architecture. (Sec. \ref{section:s_network})

\noindent(2) The details of data augmentation in training. (Sec. \ref{subsection:s_aug})

\noindent(3) The details of Chained-Anchors setting. (Sec. \ref{subsection:s_anchor})

\noindent(4) The detailed experiment results of CTracker and the qualitative comparison with other SOTA methods, including POI \cite{yu2016poi} and Tracktor \cite{bergmann2019tracking}. (Sec. \ref{section:s_exp})

\noindent(5) The experiment of adding the appearance feature to CTracker. (Sec. \ref{section:s_feature})

\section{Details of Network Architecture\label{section:s_network}}

\begin{figure*}[ht]
\vspace{-5mm}
\centering{}\includegraphics[width=0.54\columnwidth]{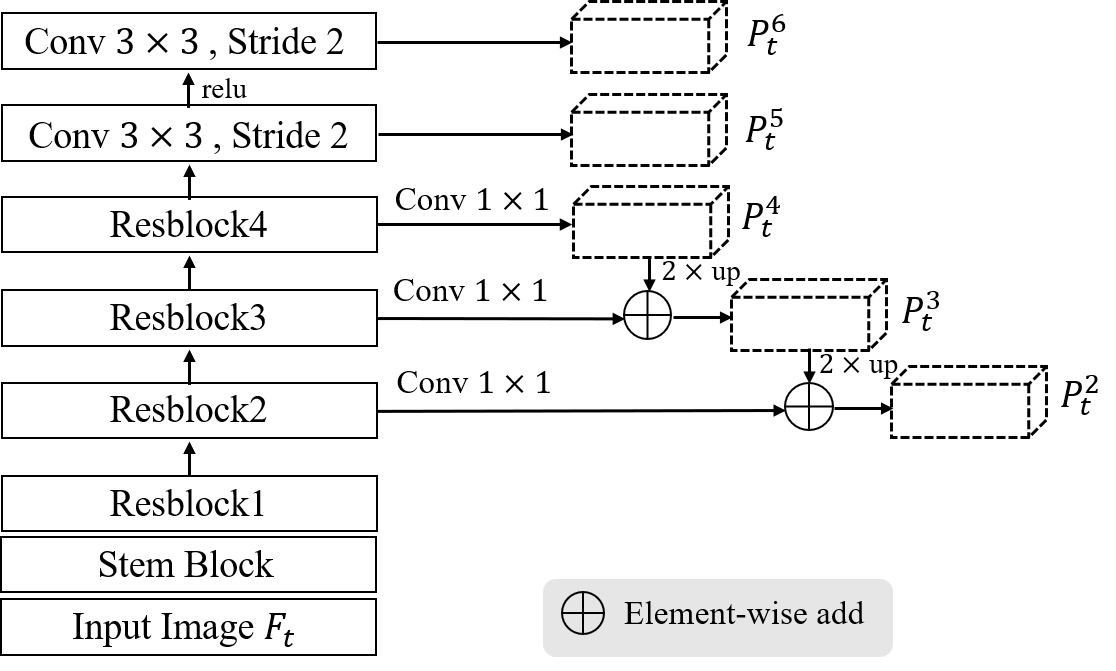}
\vspace{-5mm}
\caption{\label{fig:backbone}\textbf{The detailed architecture of backbone in CTracker network.}}
\vspace{-3mm}
\end{figure*}

As in Fig. \ref{fig:backbone}, we refer to Resnet50 \cite{he2016deep} and FPN \cite{lin2017feature} to build multi-scale feature representations at five scale levels, we denote them as $\{P_2, P_3, P_4, P_5, P_6\}$.

\begin{figure*}[ht]
\centering{}\includegraphics[width=0.90\columnwidth]{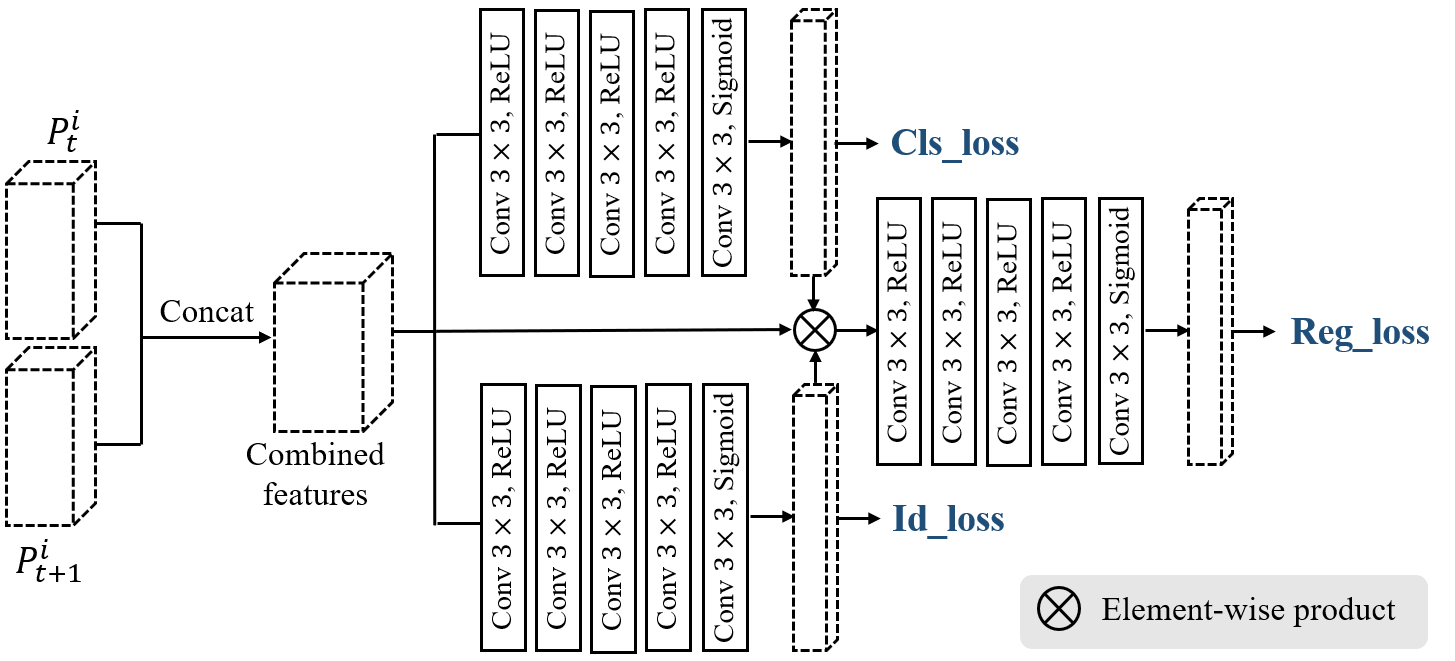}
\vspace{-3mm}
\caption{\label{fig:head}\textbf{The detailed architecture of prediction head in CTracker network.} $P_t^i$ and $P_{t+1}^i$ are the multi-scale feautres of two adjacent frames, where $i \in \{2,3,4,5,6\}$.}
\vspace{-3mm}
\end{figure*}

Then we combine the features from two adjacent frames at each scale for subsequent prediction, as in Fig. \ref{fig:head},. With the combined features, we apply two parallel branches to perform object classification and ID verification. The two branches consist of four consecutive 3$\times$3 conv layers interleaved with ReLU activations to perform feature learning for specific tasks, above which a 3$\times$3 conv with Sigmoid activation is appended to predict the confidence.  

Finally, we gather the above two predictions by multiplication to get the joint attention map. Since the attention map has the same spatial size as the combined features but with only single channel, we first apply broadcasting on the attention map so that they have compatible shapes, then we perform the attention guidance using element-wise product. With the attention-assistant features, we use the paired boxes regression branch with four conv layers to generate paired boxes for objects of interest. All the box pairs generated from the five scales are post-processed with soft-nms \cite{softnms} together. 

\section{Details of Implementation}

\subsection{Data Augmentation\label{subsection:s_aug}}
In order to construct a robust model for objects with different motion speed, we randomly sample two frames with a temporal interval of no more than 3 frames, then we reverse the order of the two frames with 50\% probability to form a training pair (\textit{i.e.}, $1<=|\delta|<=3$ in Sec. 3.4 of the main text). To further prevent over-fitting, each frame in the pair will be applied with the same data augmentations as follows:\\
(1) Randomly apply some photometric distortions introduced in SSD \cite{liu2016ssd}.\\
(2) Randomly crop a patch of the size determined by multiplying a random factor in the interval [0.3, 0.8] with the image’s shorter side. Note that we only keep those ground truths whose IoMs (Intersection over Min-area) with the cropped patch are greater than 0.2.\\
(3) With 20\% probability, expand the cropped patch using a random factor ranging in [1, 3] by padding with the mean pixel value from ImageNet.\\
(4) Flip the expanded patch randomly and resize it to a square patch with the size equivalent to the half of the original image’s shorter side.

\subsection{Chained-Anchors Setting\label{subsection:s_anchor}}
To determine the scales of Chained-Anchors, we run k-means clustering \cite{yolov2} on all ground truth bounding boxes in the dataset, then we pick five cluster centroids as the scales for Chained-Anchors in different levels of FPN. In our experiments, we use Chained-Anchors of scales $\{38, 86, 112, 156, 328\}$ for $\{P_2, P_3, P_4, P_5, P_6\}$ respectively, and the same ratio of 2.9 is taken for Chained-Anchors of all scales.

\section{Detailed Experiment Results\label{section:s_exp}}

The detailed experiment results of CTracker on MOT16 \cite{milan2016mot16} test dataset and MOT17 test dataset are displayed in Table \ref{tab:mot16_detailed} and Tabel \ref{tab:mot17_detailed}. 

\begin{table}[t!]
\vspace{-10pt}
\renewcommand\arraystretch{1.2}
\centering
\caption{\textbf{Detailed tracking results of CTracker on MOT16 test dataset}.}\label{tab:mot16_detailed}
\vspace{-10pt}
\scriptsize{
\setlength{\tabcolsep}{2.0mm}{
\begin{tabular}{|c|cccccccc|}

\hline

Sequence & MOTA$\uparrow$ & IDF1$\uparrow$ & MOTP$\uparrow$ & MT$\uparrow$ & ML$\downarrow$ & FP$\downarrow$ & FN$\downarrow$ & IDS$\downarrow$\\

\hline
MOT16-01 & 42.0 & 39.3 & 79.9 & 30.4\% & 30.4\% & 713 & 2918 & 77\\
MOT16-03 & 83.6 & 65.5 & 78.3 & 81.1\% & 0.7\% & 5600 & 11024 & 520\\
MOT16-06 & 54.7 & 52.8 & 77.1 & 27.6\% & 24.0\% & 795 & 4158 & 273\\
MOT16-07 & 52.7 & 41.4 & 78.6 & 22.2\% & 13.0\% & 587 & 6884 & 249\\
MOT16-08 & 37.2 & 35.2 & 81.8 & 19.0\% & 33.3\% & 499 & 9824 & 190\\
MOT16-12 & 46.7 & 53.5 & 78.5 & 19.8\% & 37.2\% & 112 & 4250 & 59\\
MOT16-14 & 43.7 & 43.0 & 77.1 & 12.8\% & 32.9\% & 628 & 9247 & 529\\
\hline
Total & 67.6 & 57.2 & 78.4 & 32.9\% & 23.1\% & 8934 & 48305 & 1897\\
\hline
\end{tabular}}}
\end{table}

\begin{table}[t!]
\renewcommand\arraystretch{1.2}
\centering
\caption{\textbf{Detailed tracking results of CTracker on MOT17 test dataset}.}\label{tab:mot17_detailed}
\vspace{-10pt}
\scriptsize{
\setlength{\tabcolsep}{2.0mm}{
\begin{tabular}{|c|cccccccc|}

\hline

Sequence & MOTA$\uparrow$ & IDF1$\uparrow$ & MOTP$\uparrow$ & MT$\uparrow$ & ML$\downarrow$ & FP$\downarrow$ & FN$\downarrow$ & IDS$\downarrow$\\

\hline
MOT17-01 & 51.2 & 44.4 & 78.7 & 25.0\% & 29.2\% & 202 & 2891 & 54\\
MOT17-03 & 84.9 & 66.5 & 77.9 & 83.1\% & 0.7\% & 5133 & 10211 & 479\\
MOT17-06 & 56.1 & 55.2 & 78.2 & 29.7\% & 24.3\% & 516 & 4398 & 261\\
MOT17-07 & 50.2 & 41.0 & 79.3 & 21.7\% & 23.3\% & 424 & 7761 & 228\\
MOT17-08 & 31.6 & 29.6 & 81.2 & 14.5\% & 42.1\% & 405 & 13828 & 212\\
MOT17-12 & 47.0 & 55.7 & 79.2 & 18.7\% & 35.2\% & 91 & 4432 & 69\\
MOT17-14 & 39.5 & 42.7 & 77.4 & 10.4\% & 30.5\% & 657 & 9976 & 540\\
\hline
Total & 66.6 & 57.4 & 78.2 & 32.2\% & 24.2\% & 7428 & 53497 & 1843\\
\hline
\end{tabular}}}
\end{table}

\begin{figure*}[t!]
\centering{}\includegraphics[width=0.98\columnwidth]{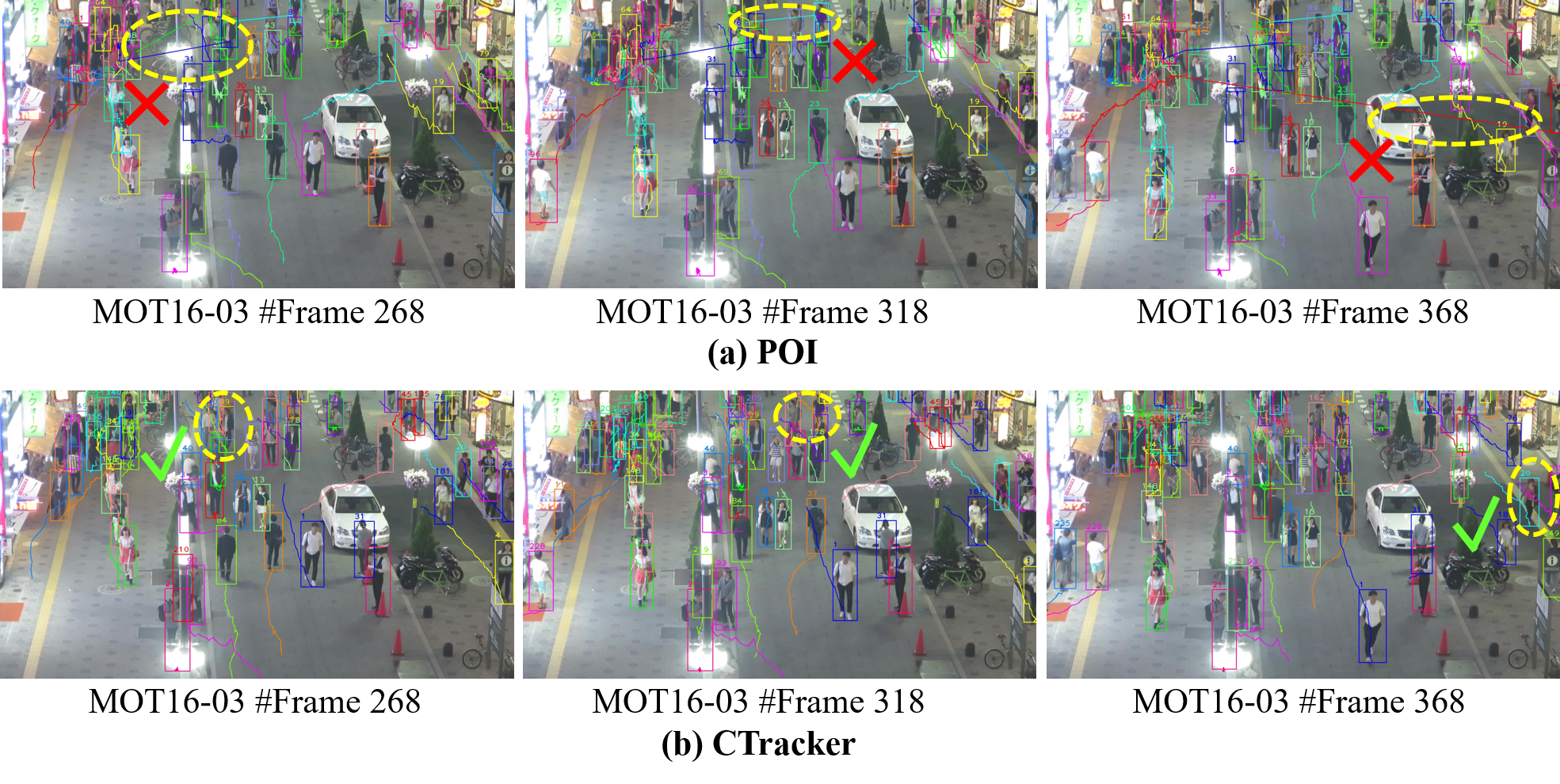}
\vspace{-5mm}
\caption{\label{fig:poi}\textbf{Qualitative comparison of POI (a) and our CTracker(b).}}
\vspace{-0.04in}
\end{figure*}

\begin{figure*}[t!]
\centering{}\includegraphics[width=0.98\columnwidth]{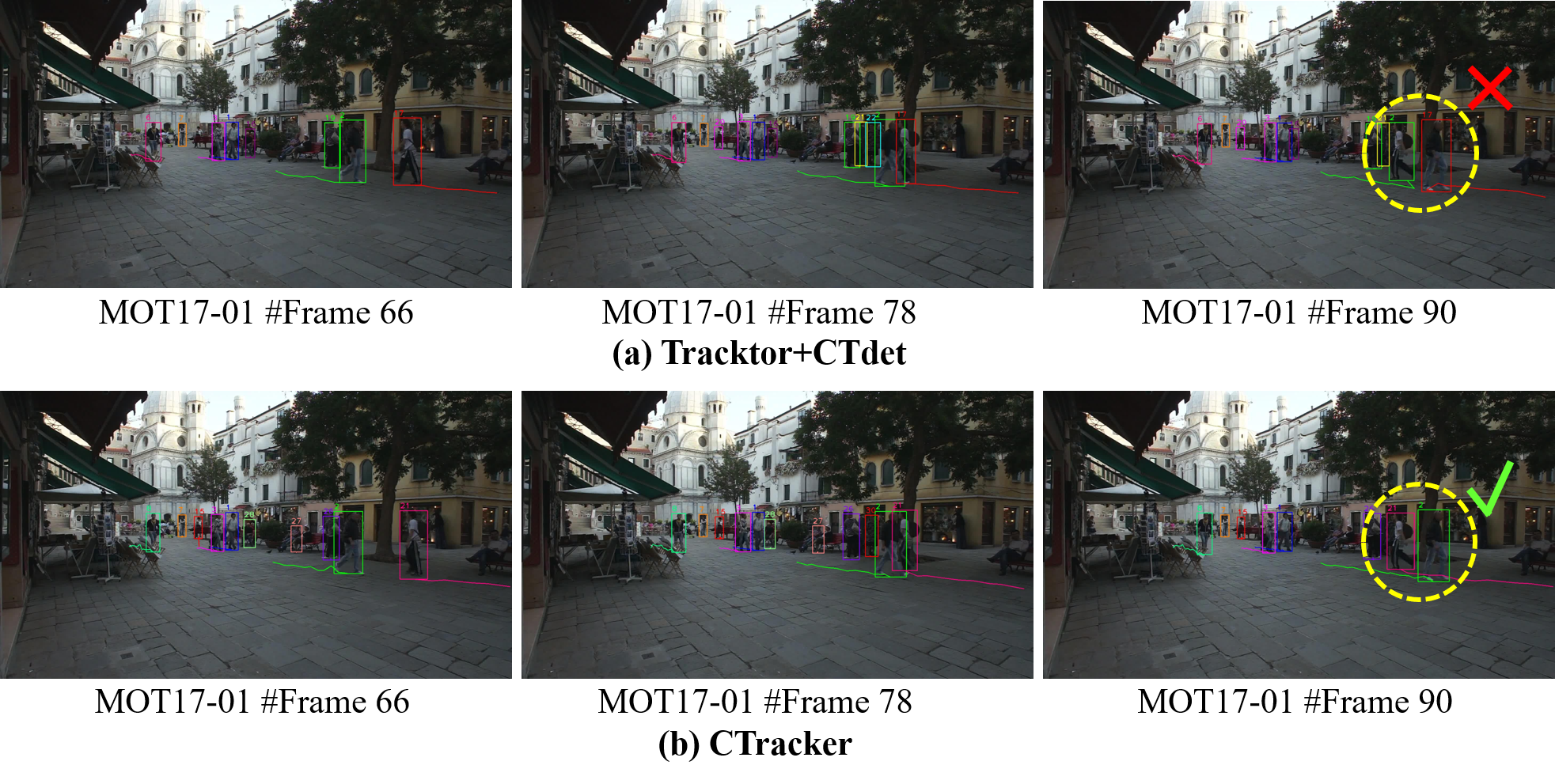}
\vspace{-5mm}
\caption{\label{fig:tracktor}\textbf{Qualitative comparison of Tracktor (a) and our CTracker(b).}}
\vspace{-0.04in}
\end{figure*}

Moreover, we select two representative qualitative cases to compare our CTracker with the private detection online SOTA method POI \cite{yu2016poi} and the public detection online SOTA method Tracktor \cite{bergmann2019tracking}. Fig. \ref{fig:poi} displays the tracking results of POI and our CTracker in sequence MOT16-03. In Fig. \ref{fig:poi}(a), using the POI method, long-term cross-frame tracking drift occurs in several trajectories, which are marked with yellow dotted circles. While in Fig. \ref{fig:poi}(b), using our CTracker method, there is no long-term cross-frame tracking drift in all the trajectories. For simplicity and efficiency, we focus on the short-term tracking based on the Chained-Anchors and abandon using the patch-level ReID features of the detected boxes like POI to enhance long-term cross-frame tracking, which may reduce some trajectory integrity to a certain degree while improve the trajectory accuracy greatly. In Fig. \ref{fig:tracktor}(a), using the Tracktor method with the same detection of our CTracker, there is a ID switch of trajectory 2 and trajectory 17 due to the occlusion, which is marked with a yellow dotted circle. While in Fig. \ref{fig:tracktor}(b), using our CTracker method, the two trajectories representing the same pedestrians are generated correctly due to the accurate box pair association in the CTracker network, which demonstrates the effectiveness of our CTracker in the hard occlusion scene. More complete and clear visualization tracking comparison is displayed in the video attachments.

\section{Appearance Feature Experiment\label{section:s_feature}}

\CRpjl{In the main text, to keep the simplicity and efficiency of our CTracker, we abandon using the patch-level ReID features of the detected boxes like other MOT methods to enhance cross-frame data association. In fact, we conduct a appearance feature experiment though we think that it is not related to our main innovations. In the node chaining module, expect for the IoU affinity, we calculate the appearance similarity by adding in the appearance features (256-dim vector from the feature map before the output convolution in the ID verification branch). On MOT16, MOTA increases from 67.6 to 68.5, IDF1 increases from 57.2 to 61.8, IDS decreases from 1897 to 983. While the tracking speed decreases from 34.4fps to 29.2fps. Therefore, We can get better tracking performance when speed loss is acceptable, demonstrating the good expandability of CTracker.}


\end{document}